\documentclass[runningheads]{llncs}
\usepackage{graphicx}
\usepackage{amsmath}
\usepackage{amssymb}
\usepackage{subcaption}
\usepackage{algorithm,algorithmic, multicol}
\usepackage{mathtools}
\usepackage{cite}
\usepackage{wrapfig}
\usepackage{sidecap}

\usepackage[absolute,overlay]{textpos}

\usepackage{xcolor}

\newcommand{\loss}{\mathcal{L}}

\newcommand{\z}{\mathbf{z}}
\newcommand{\x}{\mathbf{x}}
\newcommand{\J}{\mathbf{J}}
\newcommand{\G}{\mathbf{G}}
\DeclareMathOperator{\MF}{MF}

\graphicspath{{figures/}}

\begin{document}
\title{Fast Approximate Geodesics for Deep Generative Models}
%
%
\author{Nutan Chen \inst{1}  \and
 Francesco Ferroni\inst{2} \and 
 Alexej Klushyn\inst{1} \and
 Alexandros Paraschos\inst{1} \and
 Justin Bayer\inst{1} \and
 Patrick van der Smagt\inst{1}}

 \authorrunning{N. Chen et al.}
\institute{Machine Learning Research Lab, Volkswagen Group$^1$ \\ $\enspace \enspace \enspace \enspace$ Autonomous Intelligent Driving GmbH$^2$}

%
\maketitle              
\begin{abstract}
The length of the geodesic between two data points along a Riemannian manifold, induced by a deep generative model, yields a principled measure of similarity. 
Current approaches are limited to low-dimensional latent spaces, due to the computational complexity of solving a non-convex optimisation problem. 
We propose finding shortest paths in a finite graph of samples from the aggregate approximate posterior, that can be solved exactly, at greatly reduced runtime, and without a notable loss in quality.
Our approach, therefore, is hence applicable to high-dimensional problems, e.g., in the visual domain. 
We validate our approach empirically on a series of experiments using variational autoencoders applied to image data, including the Chair, FashionMNIST, and human movement data sets.

\end{abstract}

\begin{textblock*}{15cm}(5cm,25cm) 
28th International Conference on Artificial Neural Networks, 2019
\end{textblock*}

\section{Introduction}

Estimating the similarity between data points is central to data processing pipelines. 
In computer vision it is employed for matching points from frames \cite{ScharsteinS02} and for visual place recognition \cite{LowryS0LCCM16}, where it is used for the visual detection of places despite visual changes due to weather or lighting conditions.
For a method to be successful, certain invariances have to either be used as an inductive bias or presented through data. This necessitates for highly expressive models.

Recently, deep learning has enabled training generative models on large-scale databases, as typically found in computer vision \cite{brock2018large,kingma2018glow}. 
Such models have been used for similarity estimation from the perspective of Riemannian manifolds in the context of Gaussian process latent variable models \cite{TosiHVL14}. 
As such non-parametric approaches scale poorly with data set size, several authors \cite{ChenKK2018metrics,chen2018active,arvanitidis2017latentICLR} proposed the marriage of Riemannian manifold-based metric learning with deep generative models. 
This results in a principled measure of similarity between two members of the data distribution by relating it to the shortest path, or, geodesic between two corresponding points in latent space.
A downside of this approach is that the geodesic has to be obtained as the solution to a non-linear optimisation problem. 
Consequently, there is high computational demand and no guarantee of the quality of the solution. 
The representation of this shortest path is also not obvious, as it is continuous in nature but ultimately has to be represented in a discrete fashion.

In contrast, local and global descriptors \cite{Lowe99} do not suffer from said computational challenges. 
As their representation is a vector over real numbers, it is compactly represented as an array of floating point numbers, paving the way for efficient indexing techniques for nearest neighbour lookup \cite{BeygelzimerKL06,DBIndykMRV97}.

This work takes a step in making Riemannian manifold based approaches faster, and hence applicable in high-dimensional settings. 
We show that spanning the latent space with a discrete and finite graph allows us to apply a classic search algorithm, $\mathrm{A}^\star$, to obtain accurate approximations of the geodesic, that are superior to the previously proposed ODE and neural network based approaches in terms of computation performance, without loss in quality. 
Once the graph has been built, estimating the geodesic is bounded for any pair of points. 

We apply the proposed framework to a toy example, that of a visual pendulum, to foster intuition of the approach, as it can be easily visualised. 
The practical applicability to more challenging data sets is then illustrated for the Chair, FashionMNIST and human motion capture data sets.

\section{Related work}
\label{sec:related_work}

A wide range of approaches has been proposed for estimating the similarity between data points.
Typically, distance metrics are assumed to be given, however they often come with certain assumptions about the data.
For example instances of the Minkowski distance require the data to be invariant to rotation under the $L2$-norm.
The Mahalanobis distance is a popular choice when the data are multivariate Gaussian distributed, as it is invariant to translation and scaling.

Transforming the data can further allow applying a known metric, even when the data does not directly fulfil all the assumptions. 
In \cite{WeinbergerS09,GoldbergerRHS04}, the authors proposed the use of linear transformations for supervised learning.
To enable an accurate measurement of even more complicated data, non-linear transformations based on neural networks were introduced in \cite{SalakhutdinovH07}.
Additionally, transformations of time-series data to constant-length spaces have been proposed in \cite{BayerOS12},
which allow applying similarity measures using recurrent neural networks.

To alleviate the problem of manually specifying a distance metric, learning distances directly from data has been proposed \cite{xing2003distance, weinberger2006distance, davis2007information, kulis2013metric}. 
This is especially useful in high-dimensional spaces, where obtaining a meaningful distance metric is challenging.
Traditional metrics may not even be qualitatively meaningful, since the ratio of the distances of the nearest and farthest neighbours to a given data point is almost $1$ for a wide variety of distributions \cite{aggarwal2001surprising}.

In recent work, \cite{TosiHVL14} suggested perceiving the latent space of Gaussian process latent variable models as a Riemannian manifold, where the distance between two data points is given as the shortest path along the data manifold. Treating the latent space as a Riemannian manifold enables the use of interpolation \cite{noakes1989cubic} and trajectory-generation \cite{crouch1995dynamic} algorithms between given points, with the advantage that the observable-space trajectory remains sufficiently close to the previously used training data \cite{ChenKK2018metrics, chen2018active, arvanitidis2017latentICLR, kumar2017semi}.
The geodesics, i.e.\ the length-minimising curve given the curvature of the manifold, have been approximated by neural networks \cite{ChenKK2018metrics,chen2018active} or represented by ODEs \cite{arvanitidis2017latentICLR}.

Beside interpolation, using the geometry of the manifold has been proposed and used in other approaches based on generative models. For instance, in~\cite{flennerhag2018transferring} the authors used task manifolds for meta-learning and in \cite{grattarola2018learning} the authors developed a constant-curvature manifold prior for adversarial autoencoders used for graph data.

Similarly to our approach, ISOMAP~\cite{tenenbaum2000global}, computes the distance between two points using a graph and uses it for modelling the latent space. Our method, in contrast, computes the geodesic for trained models and, therefore, enables its use in state-of-the-art deep generative models, e.g., VAEs and GANs \cite{goodfellow2014generative}.

\section{Methods}
\label{sec:methods}

In this section, first, we provide the necessary background approaches that our
work is based on, i.e., the variational autoencoders and the associated Riemannian geometry, and, second, we provide a detailed description of our approach.

\subsection{Variational autoencoders}
\label{sec:methods-vae}
Latent-variable models (LVMs) are defined by
\begin{equation}
\label{eq:nlvm}
p(\x) = \int p(\x|\z)\, p(\z)\, \mathrm{d}\z,
\end{equation}
where the observable data $\x \in \mathbb{R}^{N_{x}}$ are represented through
latent variables $\z \in \mathbb{R}^{N_{z}}$, that are based on hidden characteristics in $\x$.  
The integral of Eq.~(\ref{eq:nlvm}) is usually intractable and has to be
approximated through sampling \cite{hastings1970monte, gelfand1990sampling} or
variational inference (VI) \cite{KingmaW13, rezende2014stochastic}. 
Using VI, the problem is rephrased as the maximisation of the evidence lower
bound (ELBO), i.e.,
\begin{equation}
\begin{aligned}
\ln p(\x) \geq
\mathbb{E}_{q(\z)} \Big{[} \ln  \frac{p(\x|\z)\, p(\z)}{q(\z)} \Big{]} =: \loss_\text{ELBO},
\end{aligned}
\end{equation}
where $p(\x|\z)$ is the likelihood, $p(\z)$ the prior, and $q(\z)$ approximates the intractable posterior.
The distribution parameters of $q(\z) = q_{\phi}(\z|\x)$ and $p(\x|\z) =
p_\theta(\x|\z)$ can be expressed by neural networks to obtain
the variational autoencoder (VAE)~\cite{KingmaW13, rezende2014stochastic}.
A tighter bound was proposed in the importance-weighted autoencoders (IWAEs)~\cite{BurdaGS15}
through importance sampling, i.e., 
\begin{equation}
\label{eq:K}
\loss_\text{IWAE} = \mathbb{E}_{\z_{1},\dots , \z_{K} \sim q_{\phi}(\z|\x)} \Big{[} 
\ln \frac{1}{K} \sum_{k=1}^{K} \frac{p_{\theta}(\x|\z_{k}) \,p_{\theta}(\z_{k})}{q_{\phi}(\z_{k}|\x)} \Big{]},
\end{equation}
where $\ln p(\x) \geq \loss_\text{IWAE} \geq \loss_\text{ELBO}$. In our
experiments we used IWAEs.

\subsection{Riemannian geometry in variational autoencoders}
\label{sec:methods-riemannian_geometry}
A Riemannian space is a differentiable manifold $M$ with an additional metric to describe its geometric properties. 
This enables assigning an inner product in the tangent space at each point
$\mathbf{\z}$ in the latent space through the corresponding metric tensor $\G
\in \mathbb{R}^{N_z  \times N_z}$, i.e.,
\begin{equation}
\langle\z', \z'\rangle_{\z} := \z'^{T}\, \G(\z)\, \z',
\end{equation}
with $\z' \in T_{\z}M$ and $\z \in M$. $T_{\z}M$ is the tangent space.
Treating the latent space of a VAE as a Riemannian manifold allows us to compute the \emph{observation space distance} of latent variables.
Assuming we have a trajectory $\gamma:[0, 1]\rightarrow\mathbb{R}^{N_{z}}$ in the Riemannian (latent) space that is transformed by a continuous function $f(\gamma(t))$ (decoder) to an $N_{x}$-dimensional Euclidean (observation) space.
The length of this trajectory in the observation space, referred to as the Riemannian distance, is defined by
\begin{equation} 
\label{eq:riemannian_distance}
L(\gamma) = \int_0^1\phi(t)\,\mathrm{d}t,\qquad \phi(t) = \sqrt{\big<\dot{\gamma}(t), \dot{\gamma}(t)\big>_{\gamma(t)}}\,,
\end{equation}
with the Riemannian velocity $\phi(t)$ and $\dot{\gamma}(t)$ denoting the
time-derivative of the trajectory.
The metric tensor is defined as $\G=\J^{T}\J$, with $\J$ as the Jacobian of the decoder.
The trajectory which minimises the Riemannian distance $L(\gamma)$ is referred to as the shortest path geodesic.
We integrate the metric tensor with $n$ equidistantly spaced sampling points
along $\gamma$ to approximate the distance, i.e., 
\begin{equation}
    \tilde L(\gamma) \approx \frac{1}{n}\sum^n_{i=1} \phi(t_i).
\end{equation}
In our approach, we use a stochastic approximation of the Jacobian, as presented in~\cite{rifai2011higher},
\begin{equation}
\J(\z) = \lim_{\sigma \rightarrow 0 }\frac{1}{\sigma} \mathbb{E} [f(\z +
\mathbf{\epsilon}) - f(\z)],
\end{equation}
with $\epsilon \sim \mathcal{N} (0, \sigma^2 I)$, to reduce the computation time.

\subsection{Graph-based geodesics}\label{sec:methods-graph_based_geodesics}
Obtaining the geodesic is a challenging task, as for minimizing
Eq.~(\ref{eq:riemannian_distance}), we need the Hessian of the decoder during the optimisation process.
Computing it is a time consuming optimisation procedure that scales poorly
with the dimensionality of the observable and the latent space, intractable for a lot of applications. 
In addition, computing the Hessian limits the selection of the neural network's 
activation function~\cite{arvanitidis2017latentICLR, ChenKK2018metrics}.
To bypass the above-mentioned hurdles we introduce a graph-based approach, where a discrete and finite graph is built in the latent space using a binary tree data structure, a k-d tree, with edge weights based on Riemannian distances.
Once the graph has been built, geodesics can be approximated by applying a
classic search algorithm, $\mathrm{A}^\star$~\cite{doran1966experiments}. Our
approach is summarized in Algorithm~\ref{algorithm}.

\noindent\textbf{Building the graph.}
The graph is structured as a k-d tree, a special case of binary space partitioning trees, where each leaf node corresponds to a k-dimensional vector.
The nodes of the graph are obtained by encoding the observable data $\mathbf{X} = \{ \x^{(1)},\dots,\x^{(N)} \}$ into their latent representations $\z^{(i)}$.
This is done by using the respective mean values of the approximate posterior $q_{\phi}(\z^{(i)}|\x^{(i)})$. 
Each node is connected by an undirected edge to its k-nearest neighbours.
The edge weights are set to Riemannian distances $L(\gamma)$, where $\gamma$ is the straight line between the related pair of nodes.

\begin{algorithm}[tb]
	\caption{Graph based geodesic}\label{algorithm}
    \vspace*{-1em}
	\begin{multicols}{2}%
	\begin{algorithmic}
		\STATE \textbf{1. Graph building process}
		\STATE Train IWAE
		\STATE Sample $n\_nodes$ nodes
		\STATE Build a graph using K-D tree
		 \FOR{$i \leftarrow$ 1 to $n\_nodes$}
		 \FOR{$k \leftarrow$ 1 to $n\_neighbours$ }
		 \STATE Index $j$ is the $k$-th neighbour
		 \IF{$edge^{(ij)}$ or $edge^{(ji)}$ empty}
		 \STATE $edge^{(ij)} = L(\gamma^{(ij)})$
		 \ENDIF
		 \ENDFOR
		 \ENDFOR
	\end{algorithmic}
	\columnbreak
	\begin{algorithmic}
		 \STATE \textbf{2. Path search process}
		 \STATE Given two points $\z^{(i)}$
		 \IF{$\z^{(i)}\notin$ nodes}
         \STATE Insert $\z^{(i)}$ into the graph in K-D tree 
		 \FOR{$k \leftarrow$ 1 to $n\_neighbours$ }
		 \STATE Index $j$ is the $k$-th neighbour
		 \STATE $edge^{(ij)} = L(\gamma^{(ij)})$
		 \ENDFOR
		 \ENDIF
		 \STATE Search geodesic using $\mathrm{A}^\star$ 
		 \RETURN path nodes
	\end{algorithmic}
	\end{multicols}
\end{algorithm}

\noindent\textbf{Approximating geodesics.}
A classic graph-traversing method to obtain the shortest path between nodes is $\mathrm{A}^\star$ search.
It is an iterative algorithm that, given a graph $\mathcal{G}$, maintains a sorted list of nodes that can be visited in the current state.
The list is typically initialised with the starting node and is being sorted according to the estimated cost of including node $n$ into the optimal path.
The estimated cost is computed by
$f(n) = g(n) + h(n),$
where $g(n)$ is the cost of the path from the starting node $n_\mathrm{start}$ to $n$ and $h(n)$ is a heuristic function that estimates the remaining cost from $n$ to the target node $n_\mathrm{target}$.

The cost function we use in our approach is the Riemannian distance between two subsequent nodes on the path, whereas the distance on the Remannian manifold with Euclidean interpolation on the latent space is used as heuristic.
$\mathrm{A}^\star$, in order to operate, requires a heuristic function that underestimates the true cost. 
It can be shown that the proposed heuristic fulfills this requirement.
The performance of the algorithm is optimal among any other similar algorithm to the number of nodes that are being expanded.
When the target node is reached, the algorithm terminates.
The result is the shortest path through the graph regarding the Riemannian distance.
This path approximates the geodesic well as shown in Section~\ref{sec:results}.

\section{Results}
\label{sec:results}

We present an empirical evaluation of our graph-based approach for
approximating geodesics in deep generative models. We compare the geodesics
to Euclidean trajectories and show that following the geodesic leads to a
smoother interpolation in the observation space. Additionally, we 
compare the graph-based approximation to a neural network (NN) based method,
proposed in~\cite{ChenKK2018metrics}, to show that our approach does
not degrade the approximation of the geodesic and scales significantly better.

In our comparisons, the NN-based method approximates the curve $\gamma$ with 
NNs, which weights are updated during the minimisation of the trajectory length
$L(\gamma)$. Euclidean interpolation is linear interpolation in the latent
space. Piecewise Euclidean interpolation uses $\mathrm{A}^\star$ search on a graph in which the edges are the distances in the latent space. The distances of Euclidean and piecewise Euclidean interpolations are computed in Riemannian manifold after interpolation.

We use the magnification factor ($\MF$) \cite{bishop1997magnification} to show the sensitivity of the generative models in 2D latent space and evaluate the approximated geodesic.
The $\MF(\z)\eqqcolon \sqrt{\det\G(\z)}  $ can be interpreted as the scaling factor when moving from the Riemannian (latent) to the Euclidean (observation) space, due to the change of variables.

\subsection{Pendulum experiment}

\begin{figure}[t]
    \centering
    \begin{subfigure}[b]{0.48\columnwidth}
        \includegraphics[width=\textwidth]{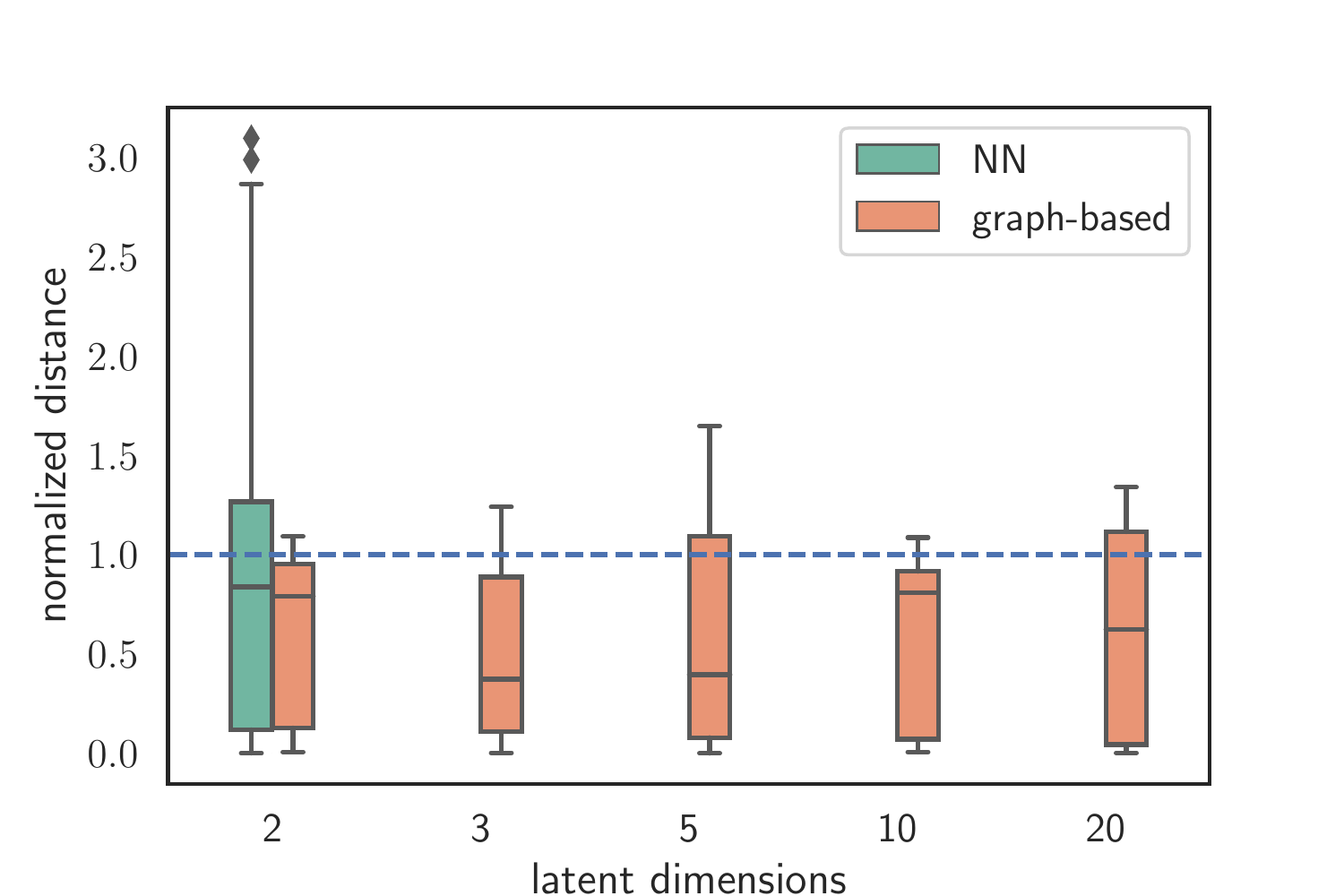}
        \caption{Normalised distance.}%
    \label{fig:method_comparision_distance}
    \end{subfigure}
    \hfill
    \begin{subfigure}[b]{0.48\columnwidth}
        \includegraphics[width=\textwidth]{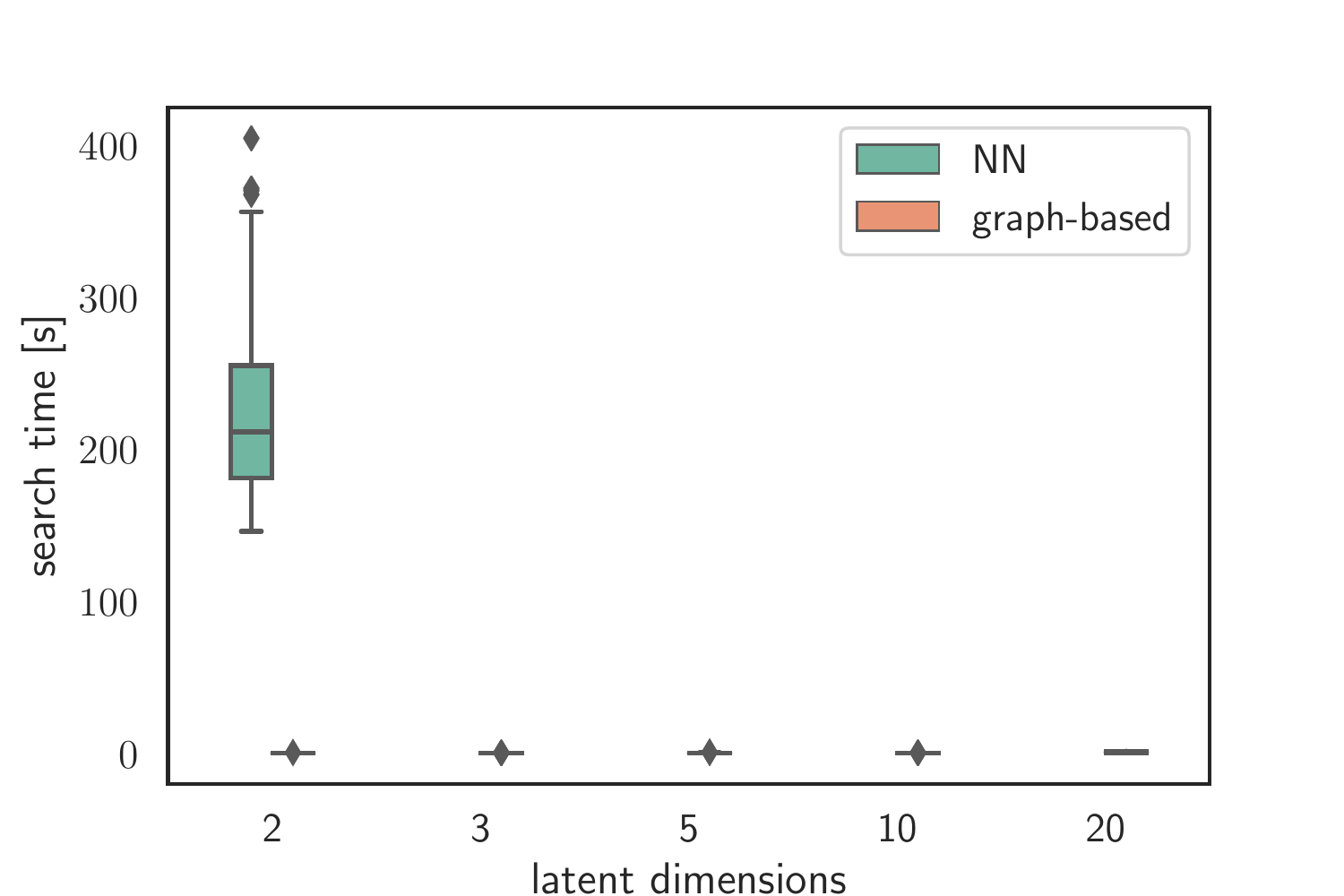}
        \caption{Path searching time.}%
    \label{fig:method_comparision_search_time}
    \end{subfigure}
    \caption{Box plot of distances and searching time using $100$ pairs of
        randomly selected data points. The box plot illustrates the median, as well as [25, 75] and [5, 95] percentiles.
        Our approach produces shorter distances and scales well to higher latent
        dimensions.
        (a) The geodesic distances are normalised to enable comparison across
        different generative models.
        The normalised distance is computed by $d_\textrm{norm}=d_\textrm{Geod.}/\textrm{mean}(d_\textrm{Eucl.})$. 
        (b) The mean of the graph-based  $\mathrm{A}^\star$ searching time is
        $0.09s$.}
    \label{fig:pendulum_ND}
\end{figure}

\begin{figure}[t]
    \centering
    \begin{subfigure}[c]{0.45\columnwidth}
        \includegraphics[width=\columnwidth]{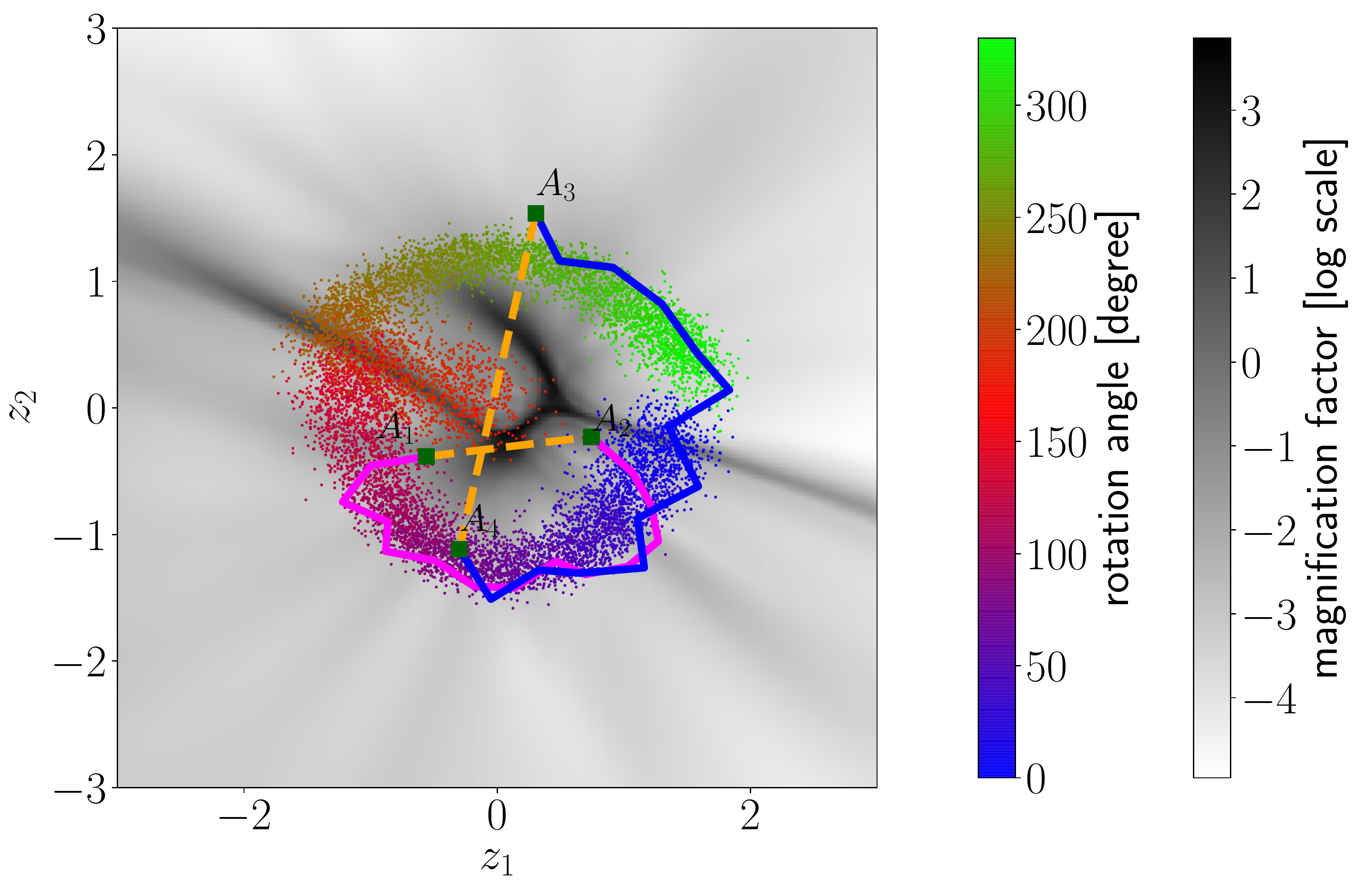}
    \end{subfigure}
    \begin{subfigure}[c]{0.54\columnwidth}
        \includegraphics[width=\textwidth]{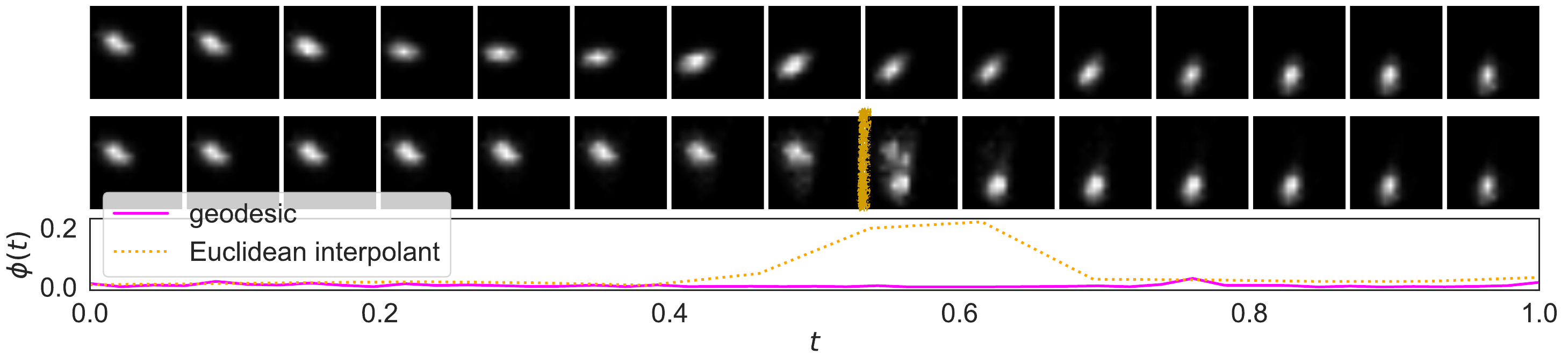}
        \includegraphics[width=\textwidth]{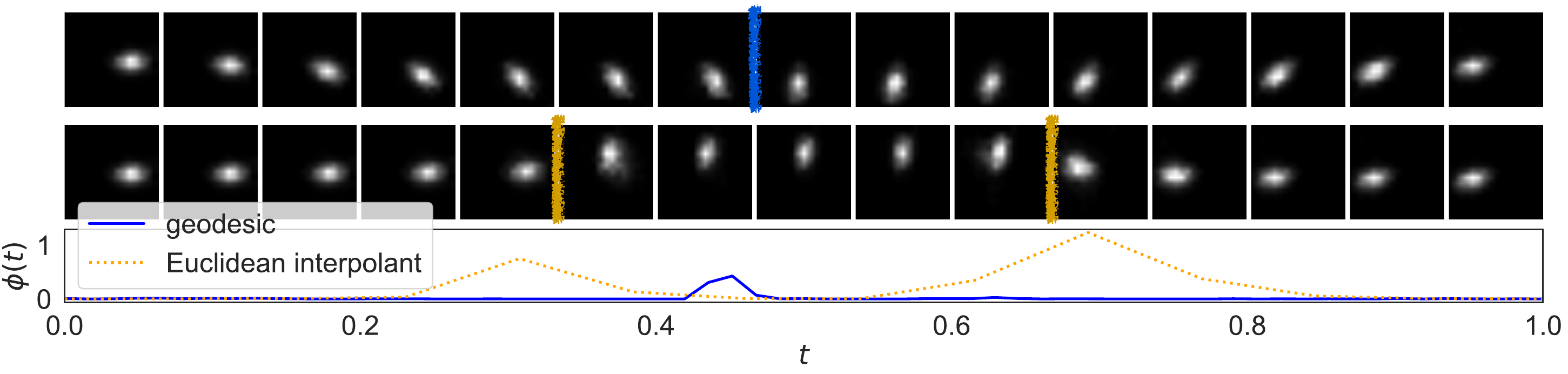}
    \end{subfigure}
    \caption{(left)
        Latent space and MF of pendulum in two dimensions. With blue and magenta
        we show the approximate geodesics and with orange the Euclidean
        interpolations.(right) The respective reconstructed images. The upper
        images series is the reconstruction using the geodesic and the lower the
        Euclidean interpolation. The geodesic reconstruction is significantly
        smoother, as we can also see from the velocity $\phi$.}
    \label{fig:pendulum_recon}
\end{figure}

The pendulum dataset contains $16 \times 16$-pixel images of a simulated
pendulum and has $T=15 \cdot10^3$ images for two different joint angle ranges, 
$R_1 = [0, 150)$ and $R_2 = [180, 330)$ degrees. To avoid overfitting, we augmented 
the dataset by adding $0.05$ Gaussian noise to each pixel.
We present our results using $\{2, 3, 5, 10, 20\}$ latent dimensions for the
IWAE and used 15 samples for the importance weighting step. 
After training, $1000$ points were chosen to build the graph. 
Each node had four nearest neighbours based on the distance in the latent space.
We generated $100$ random pairs of data points, as shown in
Fig.~\ref{fig:pendulum_ND}, for computing the distances and search time. 
We show that with the increasing latent dimensionality, the search time does not
increase, as it is dependent solely on the number of nodes. Comparing to 
\cite{chen2018active} and  \cite{arvanitidis2017latentICLR}, our approach 
does not require second-order derivatives and is significantly faster. The 
\cite{arvanitidis2017latentICLR} approach takes more time than the NN-based method, 
so we only used latter for this comparison.

Two of the generated geodesic and Euclidean interpolant trajectories, in the
case of two latent dimensions are illustrated in Fig.~\ref{fig:pendulum_recon}.
Our approach finds a trajectory in the latent space which is significantly longer than a simple 
Euclidean interpolation, but significantly smoother as it does not cross a
region of large MF. The effects of crossing such a region are shown in
Fig.~\ref{fig:pendulum_recon}(right).

\subsection{Fashion MNIST}

\begin{SCfigure}[][t]
    \centering
        \includegraphics[width=0.5\columnwidth]{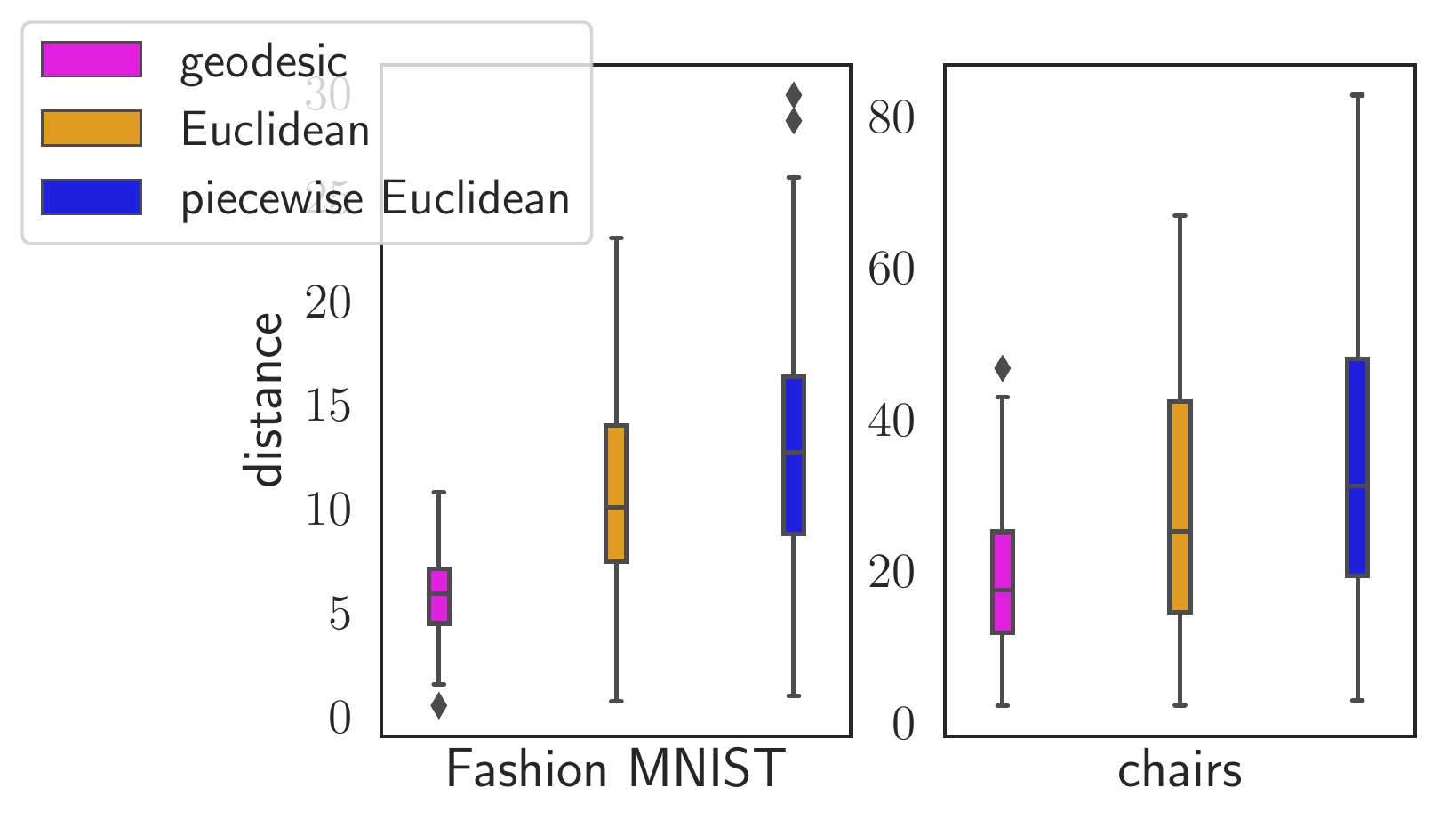}
        \caption{The distribution of distances in Fashion MNIST and chairs from
            100 randomly sampled trajectories based on geodesic, Euclidean
            interpolation or piecewise Euclidean interpolation. Both of the
            datasets are with 20D latent space.}%
        \label{fig:distance_chair_fashion}
\end{SCfigure}

\begin{figure}[t]
    \centering
    \hfill
    \begin{subfigure}[b]{0.45\textwidth}
        \includegraphics[width=\columnwidth]{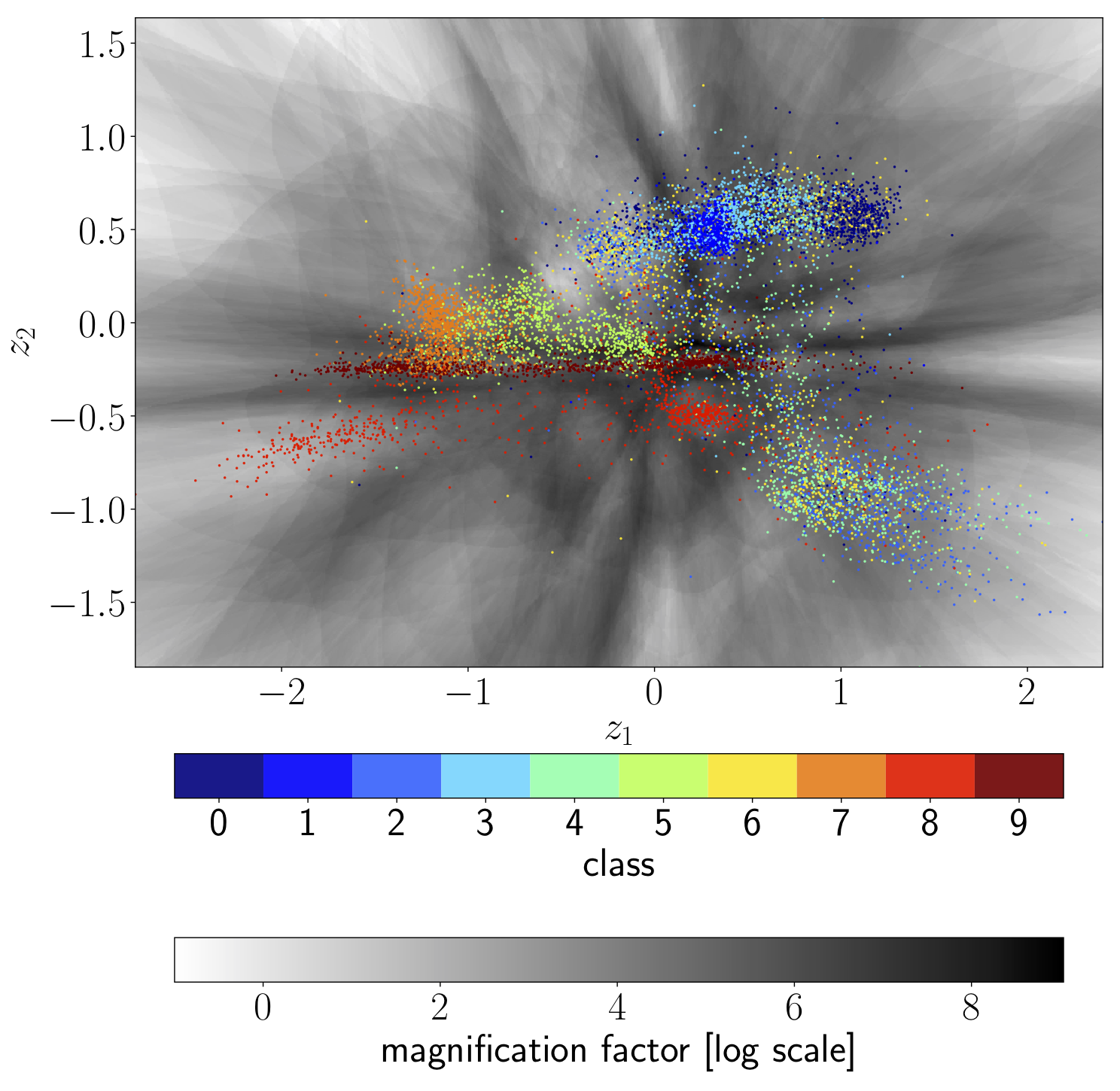}%
    \end{subfigure}
    \hfill
    \begin{subfigure}[b]{0.45\textwidth}
        \includegraphics[width=\columnwidth]{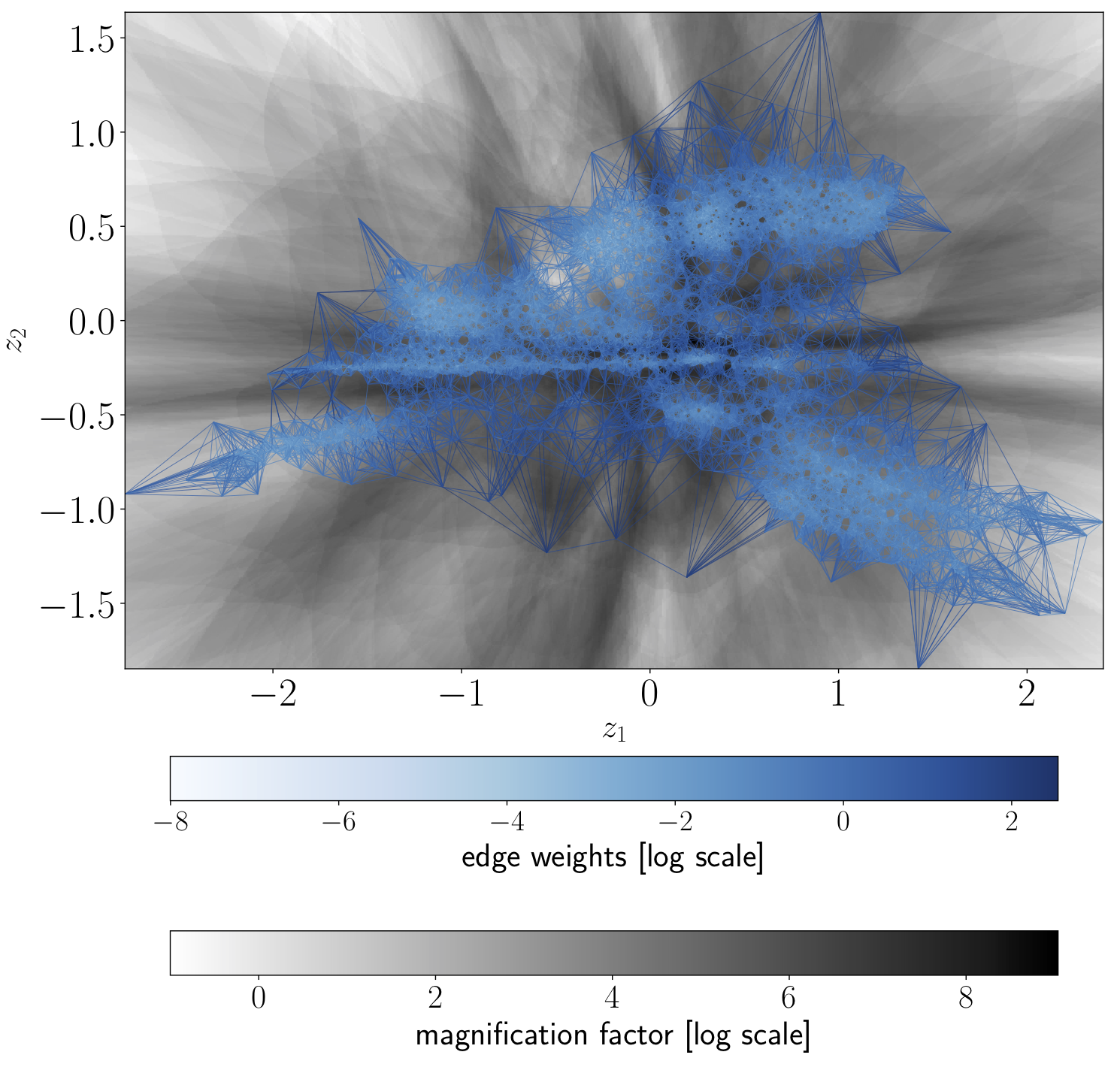}%
    \end{subfigure}
    \hfill
    \caption{(left) MF of Fashion MNIST using a 2D latent space. Points are
        sampled from the validation dataset. 
        (right) Respective graph of Fashion MNIST. The edges are weighted by
        the geodesic distance. Darker color signifies a transition with a higher
        $\MF$.}%
    \label{fig:fashionrotate_graph}
\end{figure}

\begin{figure}[t]
    \centering
    \begin{subfigure}[b]{0.49\columnwidth}
        \includegraphics[width=\textwidth]{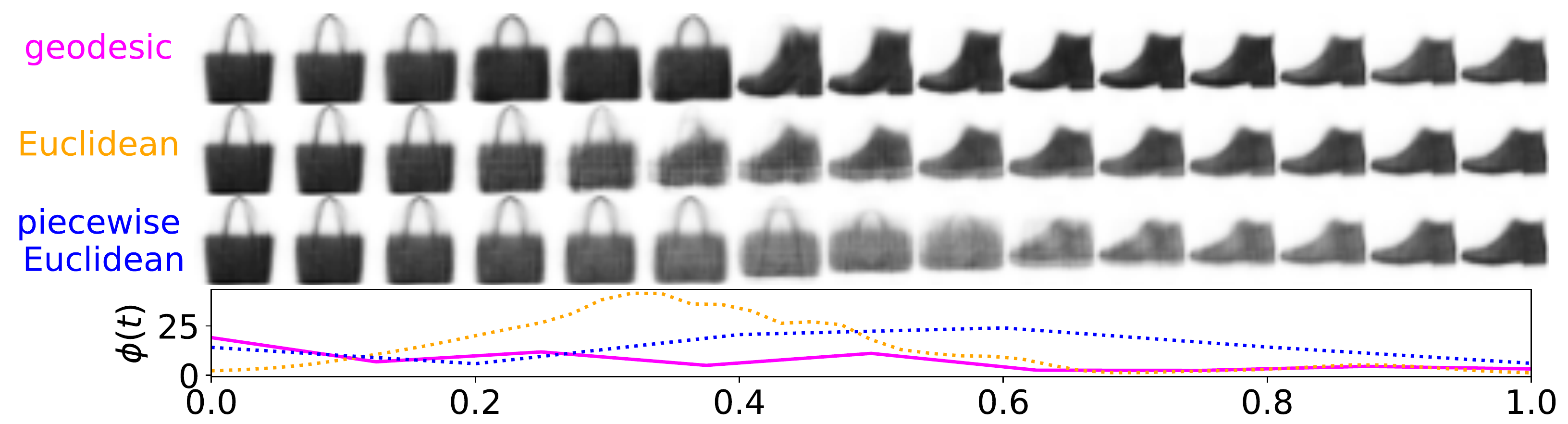}
        \includegraphics[width=\textwidth]{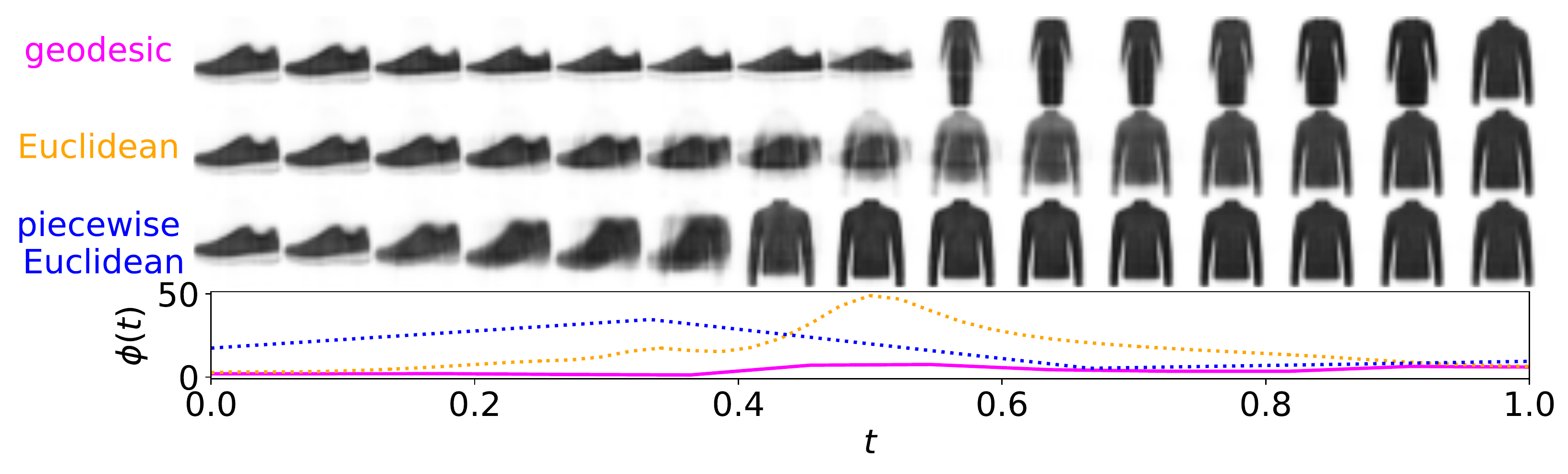}
    \end{subfigure}
    \hfill
    \begin{subfigure}[b]{0.49\columnwidth}
        \includegraphics[width=\textwidth]{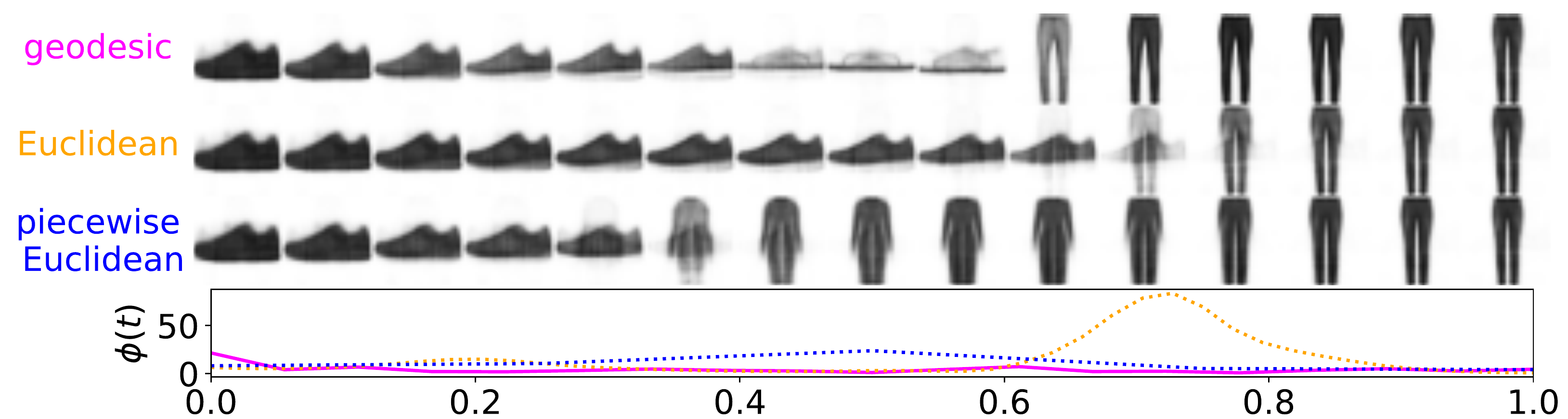}
        \includegraphics[width=\textwidth]{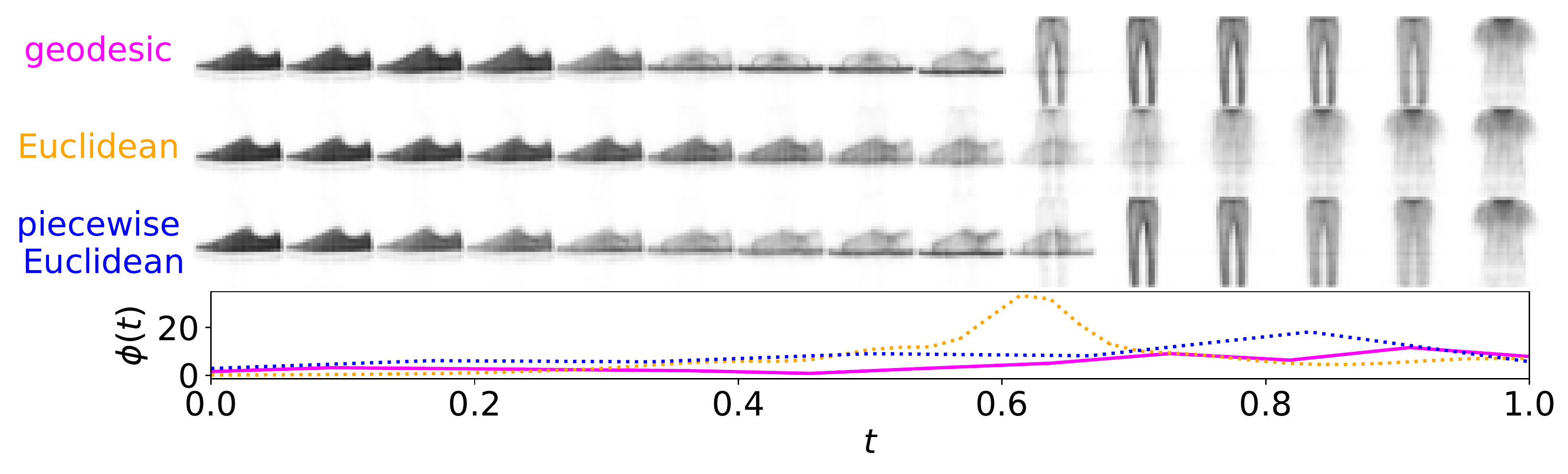}
    \end{subfigure}
    \caption{Reconstruction of Fashion MNIST with $20$ latent dimensions. 
        The geodesic outperforms Euclidean interpolation by producing interpolations 
        that visually stay on the manifold and the objects are recognisable.}
    \label{fig:fashion_recon}
\end{figure}

For the Fashion MNIST~\cite{xiao2017online} evaluation, we used the standard
training set, i.e. $28 \times 28$ pixel images, for fitting an IWAE consisting
of eight 128-neuron layers with ReLU activation~\cite{glorot2011deep}. We use
$20$ latent dimensions for our evaluations. Additionally, standard augmentation
strategies were used, e.g., horizontal flipping and jitter, and 20\% dropout, to
avoid overfitting. To generate the nodes of the geodesic graph, we randomly selected 2000 validation data points, encoded into the latent space. For each node, we selected the 20 nearest data samples, based on a Euclidean distance. For each edge we calculated the velocity and geodesic of the trajectory, using fifteen interpolation points. 

As shown in Fig.~\ref{fig:distance_chair_fashion}, we sampled $100$ trajectories between data points and calculated the geodesic distance and a distance based on Euclidean interpolation. The search time of $100$ trajectories between data points is $0.018\ s \pm 0.010$ (mean $\pm$ STD). The Euclidean-based trajectories result in consistently higher MF values, for all latent dimensions used. Therefore, the Euclidean-based trajectories cross more areas with high MF, resulting in fewer smooth transitions in the observational space. However, following the geodesic, smoothness is not guaranteed. There are cases where it is unavoidable for certain trajectories to cross an area of high magnification factor. Rather, following the geodesic can be interpreted  as the minimisation of the overall Riemannian distance for a trajectory and it is  expected that the MF will be lower than a simple Euclidean interpolation. Although piecewise Euclidean interpolation mainly follows the data manifold, it still cannot detect the high MF values; therefore, it is reasonable higher than Euclidean interpolation. Fig.~\ref{fig:fashionrotate_graph} demonstrates this property on a 2D manifold, where the edges between data samples are lighter for lower magnification factors. The areas where the MF is high, the edges are darker even if the samples are adjacent in the observational space. In such situations, the graph-based approach will therefore produce a more complex trajectory, but with a lower MF.

In Fig.~\ref{fig:fashion_recon} we present visual examples of such trajectories.
The image reconstructions are produced from the decoder of the VAE by moving
through the latent space along the trajectory specified by either the geodesic
or Euclidean interpolation. The geodesic produces images that are almost always 
recognisable, despite transiting over different classes.

\begin{figure}[t]
    \centering
    \begin{subfigure}[b]{0.49\columnwidth}
        \includegraphics[width=\textwidth]{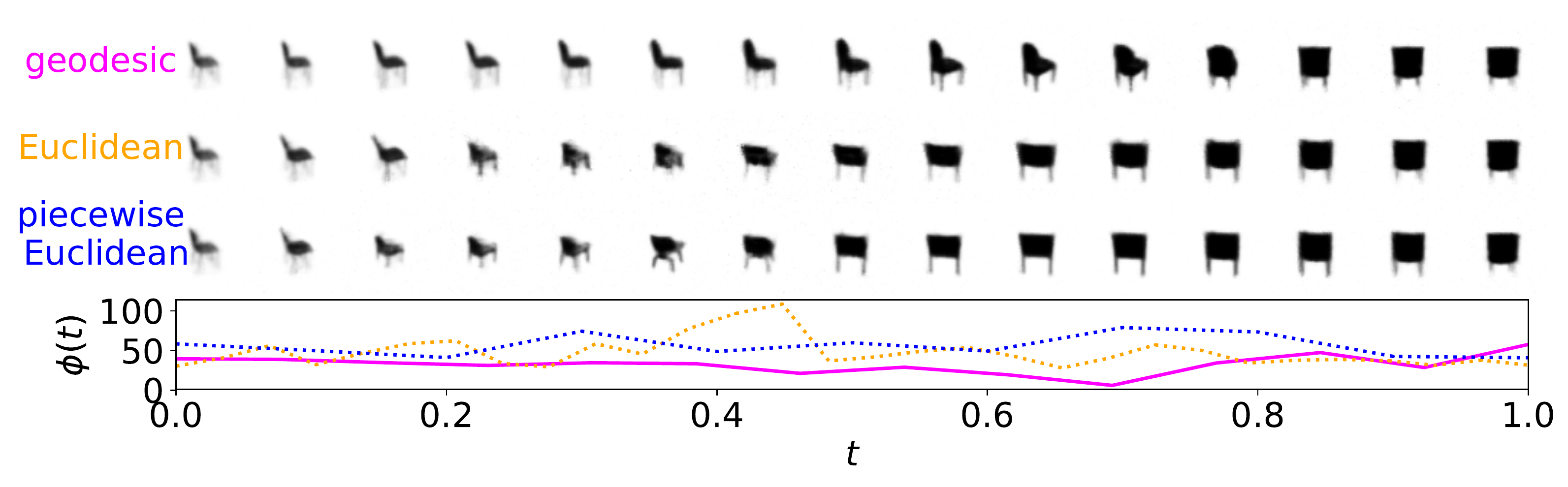}
        \includegraphics[width=\textwidth]{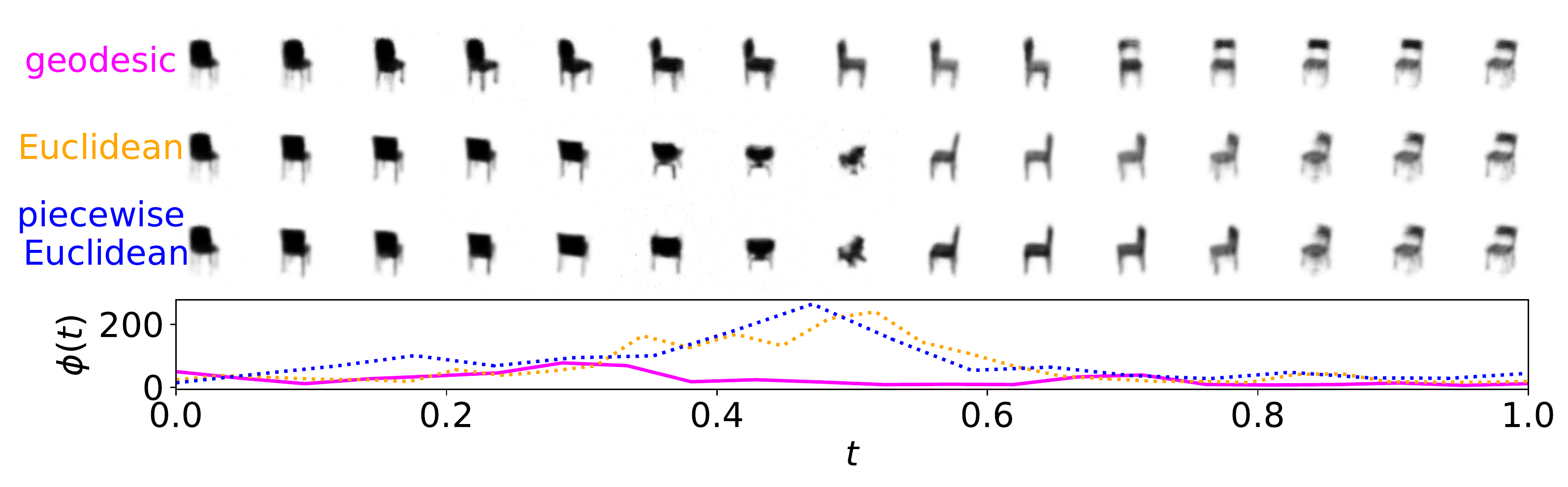}
    \end{subfigure}
    \hfill
    \begin{subfigure}[b]{0.49\columnwidth}
         \includegraphics[width=\textwidth]{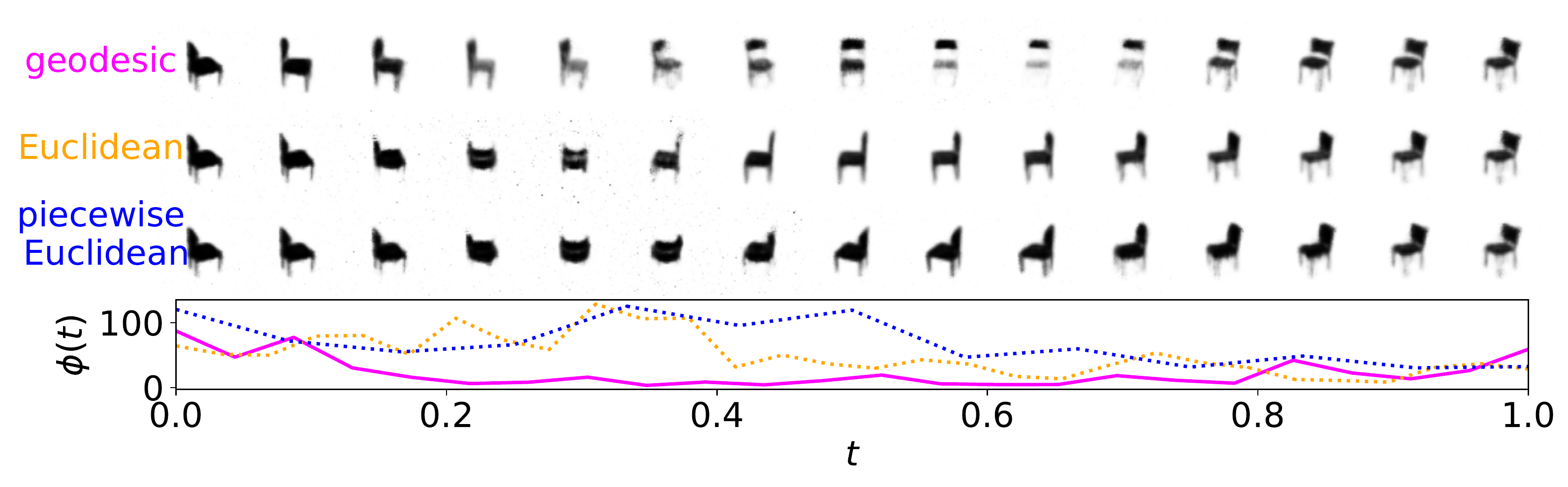}
          \includegraphics[width=\textwidth]{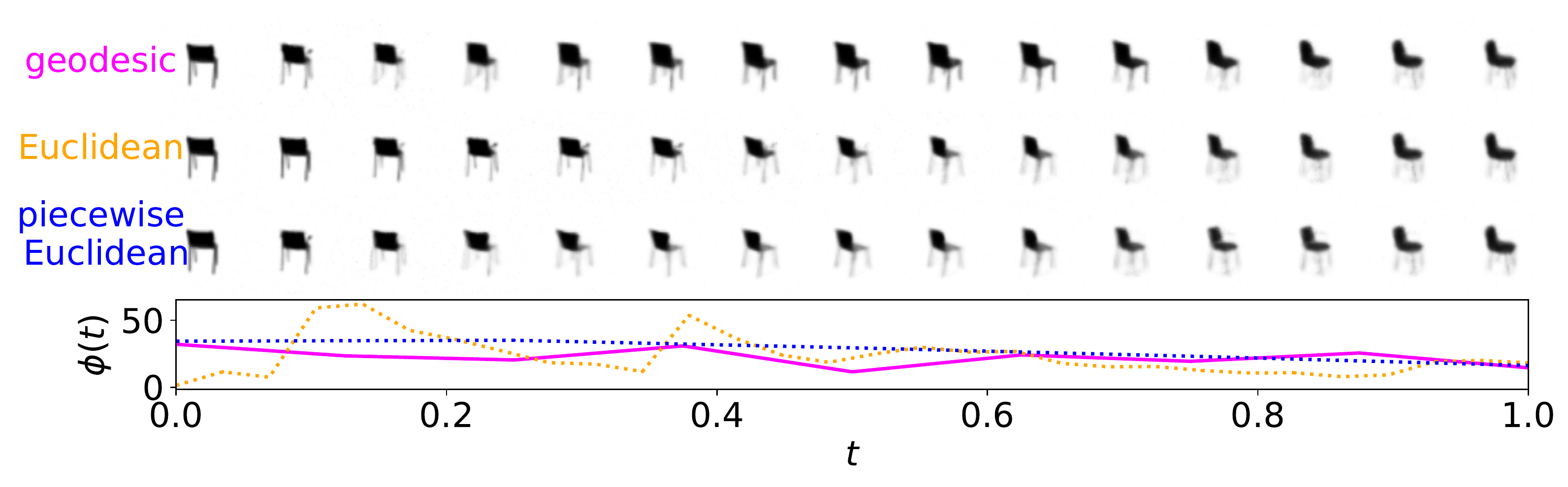}
    \end{subfigure}
    \caption{Reconstruction of Chairs dataset, with 20 latent dimensions. The
        geodesic produces a smoother interpolation in the observation space, as
        we can observe by both the image sequences and the velocity $\phi$.}%
    \label{fig:chairs_recon}
\end{figure}

\subsection{Chairs}

\begin{figure}[t]
    \centering
    \begin{subfigure}[b]{0.32\columnwidth}
        \includegraphics[width=\textwidth]{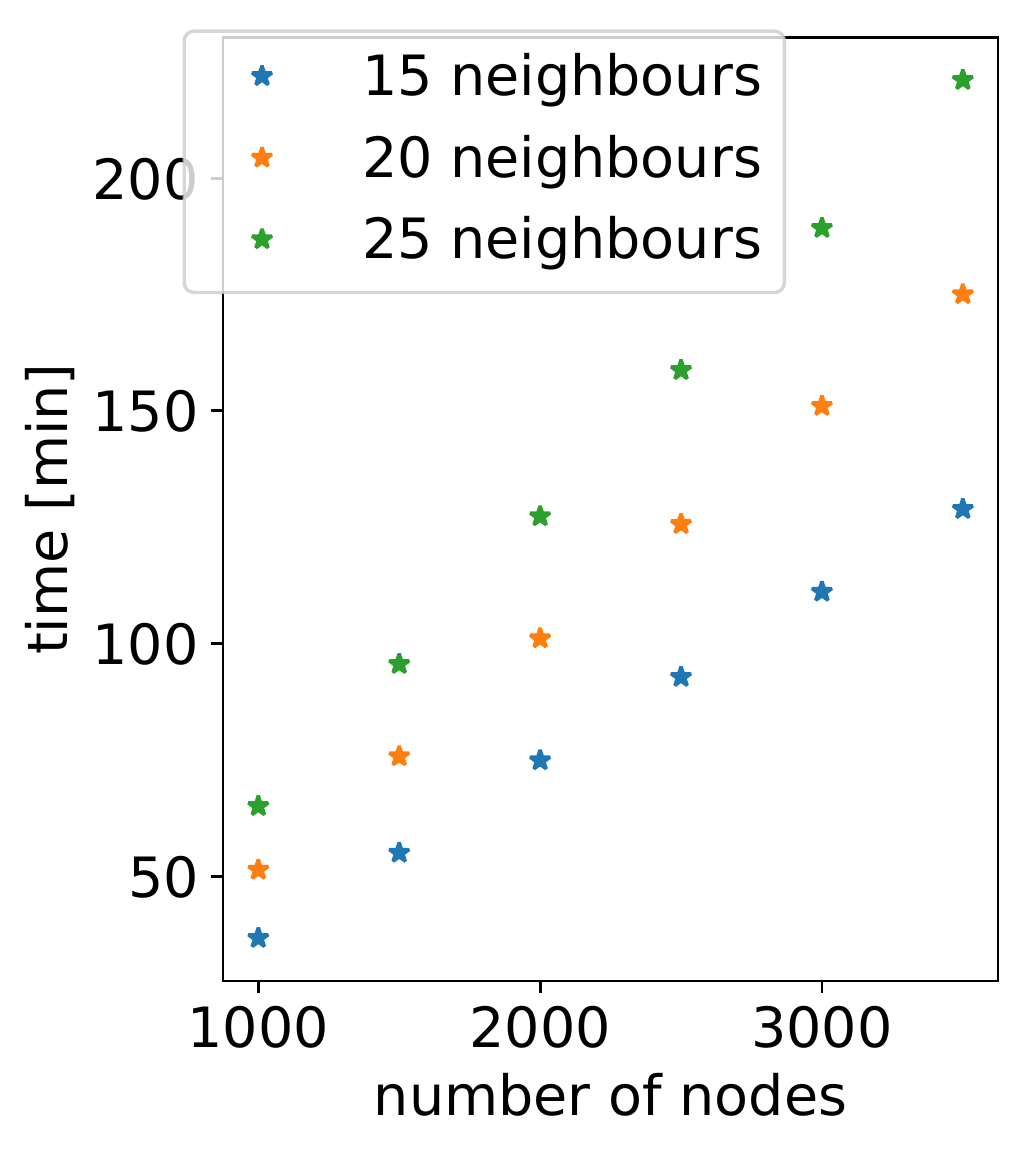}
        \caption{Graph building time}%
    \label{fig:build_graph_time}
    \end{subfigure}
    \hfill
    \begin{subfigure}[b]{0.32\columnwidth}
        \includegraphics[width=\textwidth]{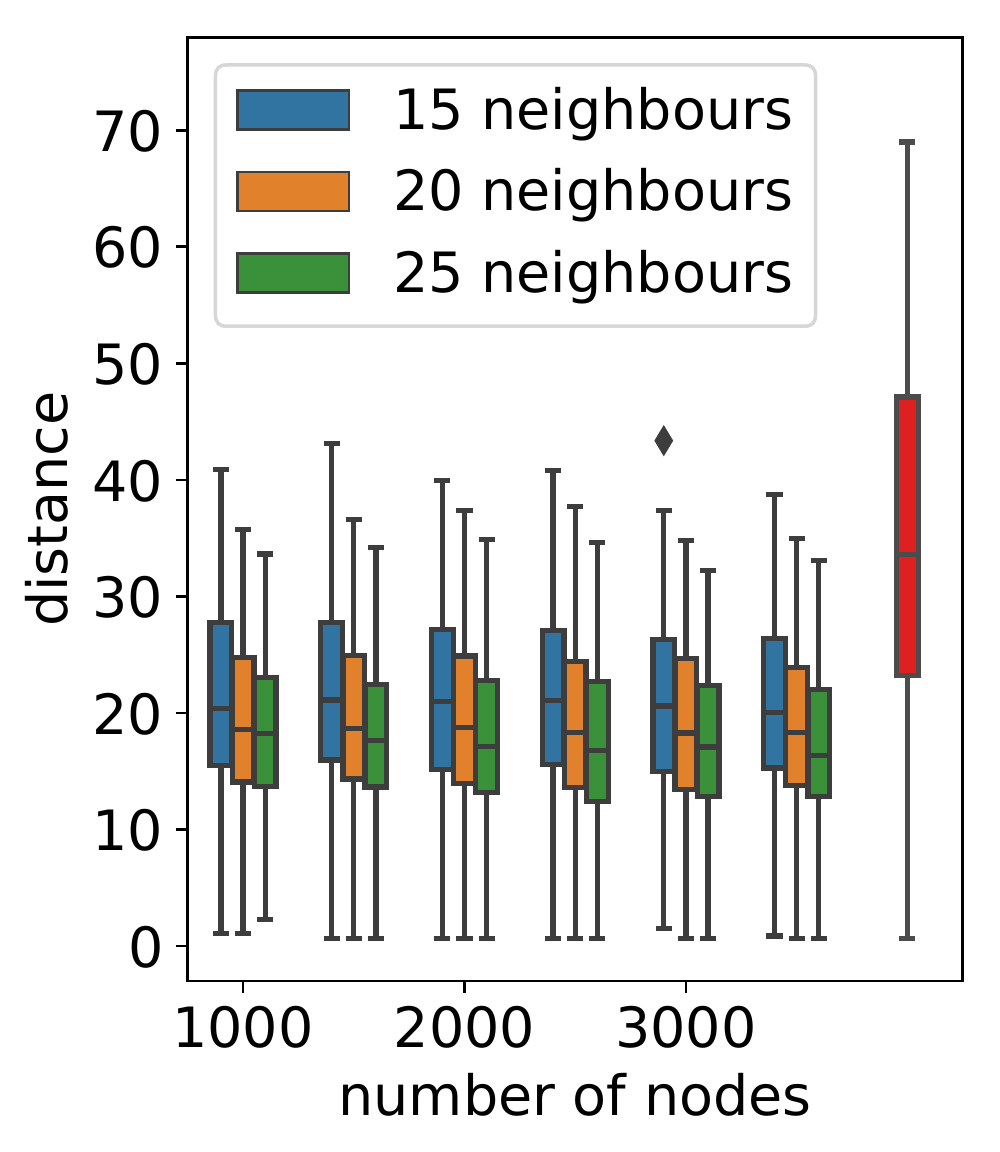}
        \caption{Distance}%
    \label{fig:build_graph_distance}
    \end{subfigure}
    \hfill
    \begin{subfigure}[b]{0.32\columnwidth}
        \includegraphics[width=\textwidth]{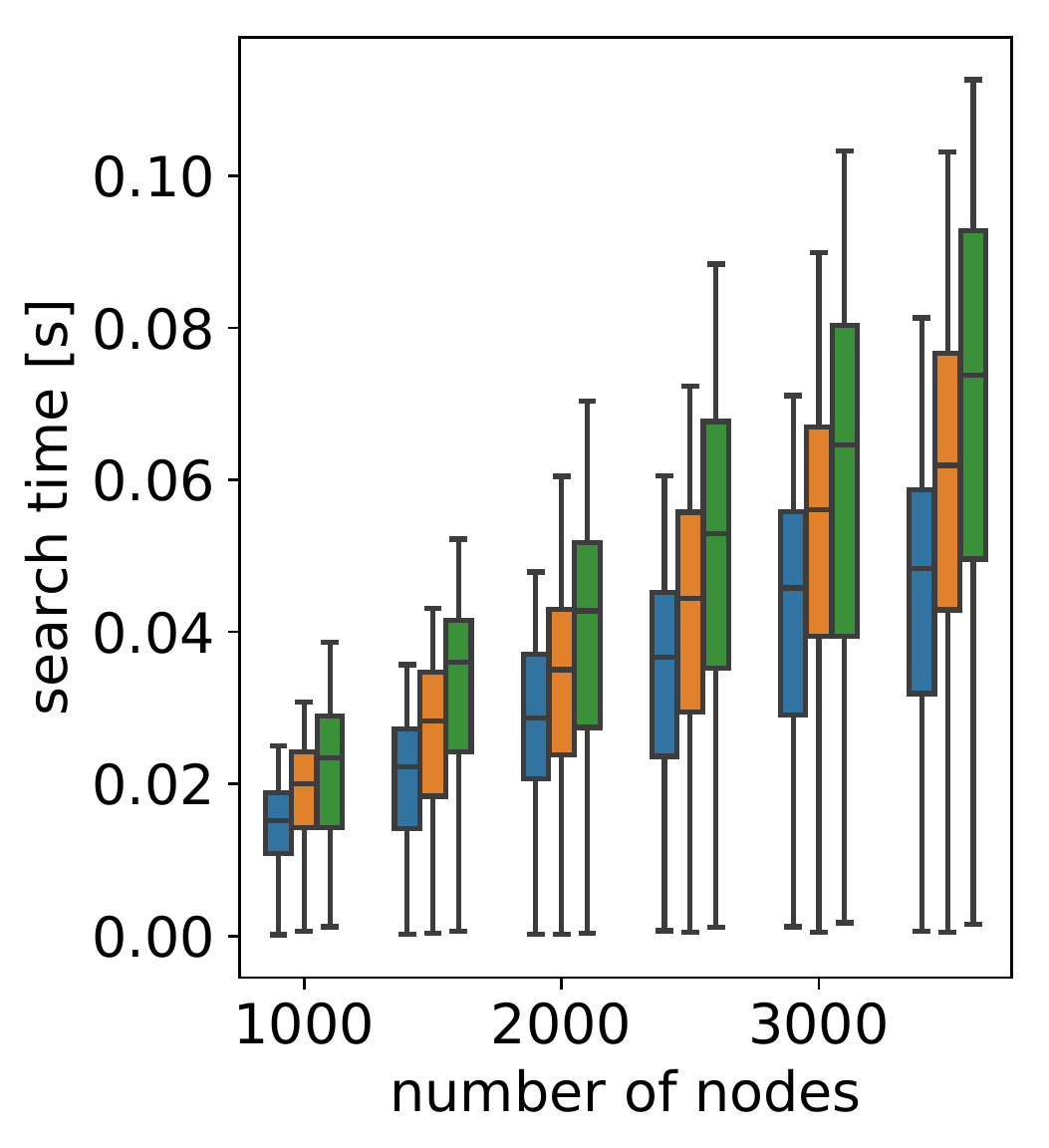}
        \caption{Search time}%
    \label{fig:build_graph_search_time}
    \end{subfigure}
    \caption{(a) The effects of the graph architectures for 20 latent dimensional latent space of the Chair dataset. 
        (b) Geodesic distances using a different amount of neighbours in
        comparison to the Euclidean distance shown in red.
        (c) Search time averaged over 100 interpolations for a different number
        of neighbours.}
    \label{fig:timing}
\end{figure}

For the chairs dataset~\cite{Aubry14}, we split chair sequences 80/20 for training and validation, 
using 74400 and 11966 images respectively. The zoom factor was set to $1.3$ and
the images rescaled to $64 \times 64$ pixels. We generated the geodesic graph 
in exactly as for Fashion MNIST.
In Fig.~\ref{fig:timing} we present a comparison of the graph building time, the average distance of
the geodesic, and the trajectory search time to the number nodes and neighbors.
Our approach scales well to the increase of the number of nodes and neighbors.

Additionally, we compare the performance of our approach to Euclidean-based
interpolation in Fig.~\ref{fig:distance_chair_fashion}, using 2000 nodes and 20
neighbors. Our approach outperforms the Euclidean-based, producing consistently
lower MF values. Reconstructions are shown in Fig.~\ref{fig:chairs_recon}.

\label{app:graph_hyper}

\subsection{Human Motions}

\begin{figure}[t]
    \centering
        \includegraphics[width=0.49\textwidth]{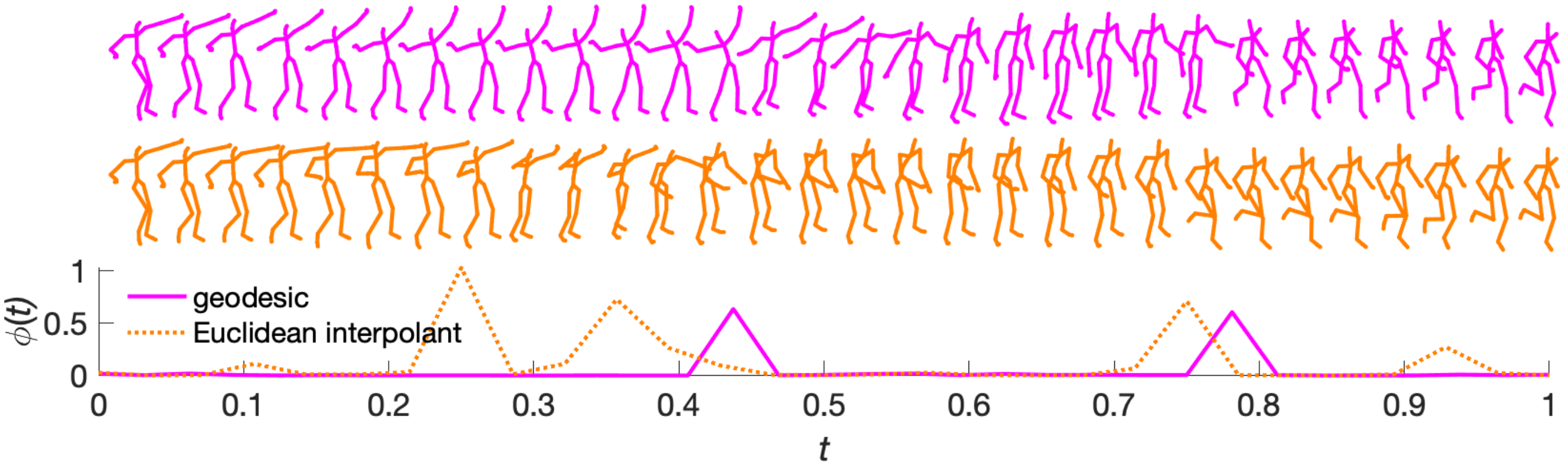} 
        \includegraphics[width=0.49\textwidth]{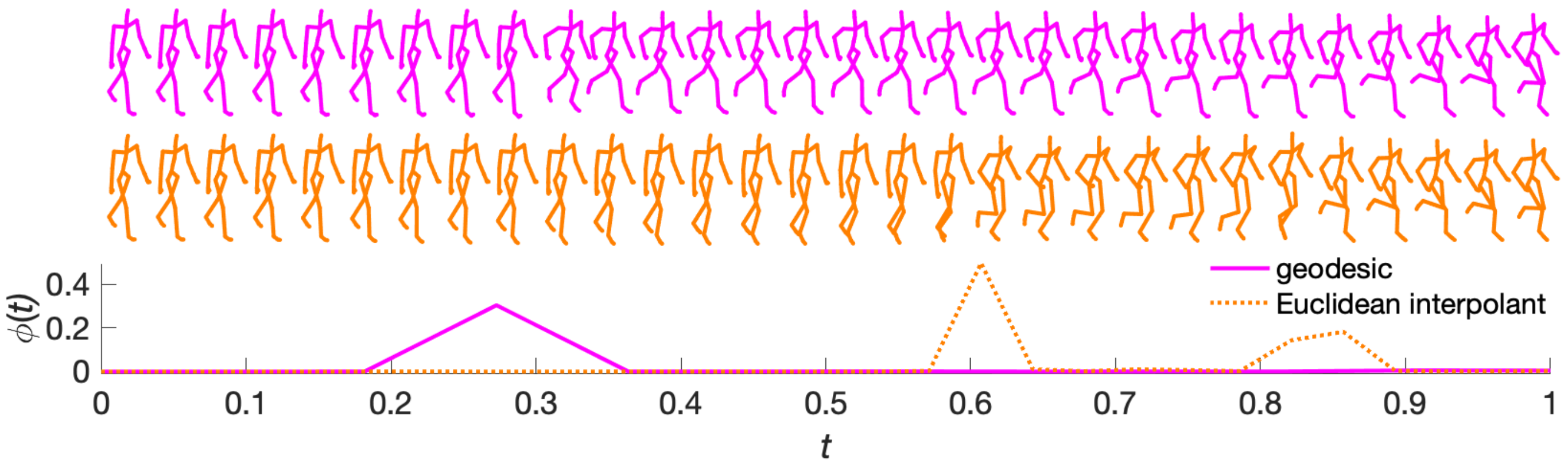} \\
        \caption{Interpolation of human motions using three latent dimensions.
            Our approach produces a smoother interpolation in comparison to the
            Euclidean.}
    \label{fig:human}
\end{figure}

We evaluate our approach in a different domain, i.e., the 
CMU human motion dataset\footnote{http://mocap.cs.cmu.edu/}, that 
includes various movements. We selected walking (subject 35), jogging (subject 35), 
balancing (subject 49), punching (subject 143) and waving (subject 143).
The input data is a 50-dimensional vector of the joint angles. In
Fig.~\ref{fig:human} we present the results using three latent dimensions. 
We observe that the Euclidean interpolation generates
trajectories crossing high MF regions. 
However, using the geodesic we are able to find similar
gestures between the classes and, subsequently, generate smoother
interpolations.

\section{Conclusion and future work}
In this paper, we demonstrate how the major computational demand of applying Riemannian geometry to deep generative models, can be sidestepped by solving a related graph-based shortest path problem instead.
Although our approach is only approximate, in our experiments on a wide variety of data sets show little loss in quality of interpolations while linear paths are consistently outperformed.
The machinery paves the way towards the application of Riemannian geometry to more applied problems that require the expressivity of deep generative models and the robustness of distance based approaches.

Further research in this topic can be how to efficiently choose the nodes such as using sigma points of unscented transform. Additionally, we use equidistance on the hidden space between two nodes for reconstruction. However, if we use equidistance on the Riemannian manifold instead, we might further improve our approach.

{\small
\bibliographystyle{ieee}
\bibliography{graph_based_geodesic.bib}
}

\end{document}